\pdfoutput=1

\documentclass[11pt]{article}

\usepackage[preprint]{acl}

\usepackage{times}
\usepackage{latexsym}
\usepackage{booktabs}
\usepackage[T1]{fontenc}

\usepackage[utf8]{inputenc}

\usepackage{microtype}

\usepackage{inconsolata}

\usepackage{graphicx}
\usepackage{algorithm}
\usepackage{algpseudocode}
\usepackage{amsmath}
\usepackage{subcaption}
\usepackage{amsmath}
\usepackage{makecell}

\title{Token-Driven GammaTune: Adaptive Calibration for Enhanced Speculative Decoding}

\author{Aayush Gautam$^*$, Susav Shrestha\thanks{Equal Contribution.}, Narasimha Reddy \\
        Department of Electrical and Computer Engineering\\
        Texas A\&M University, College Station, TX\\
        \normalsize{\texttt{\{aayushgautam, sls7161, reddy\}@tamu.edu}}}

\begin{document}
\maketitle
\begin{abstract}
Speculative decoding accelerates large language model (LLM) inference by using a smaller draft model to propose tokens, which are then verified by a larger target model. However, selecting an optimal speculation length is critical for maximizing speedup while minimizing wasted computation.  We introduce \textit{GammaTune} and \textit{GammaTune+}, training-free adaptive algorithms that dynamically adjust speculation length based on token acceptance rates using a heuristic-based switching mechanism. Evaluated on SpecBench across multiple tasks and model pairs, our method outperforms other heuristic-based approaches and fixed-length speculative decoding, achieving an average speedup of 15\% ($\pm$5\%) with \textit{GammaTune} and 16\% ($\pm$3\%) with \textit{GammaTune+}, while reducing performance variance. This makes \textit{GammaTune} a robust and efficient solution for real-world deployment.

\end{abstract}

\section{Introduction}

Large language models (LLMs) have become integral to various NLP domains, including information retrieval, conversational AI, and document summarization, driving advancements in language understanding and generation \cite{BERT, ESPN, ssd_espn}. As these applications demand real-time responsiveness and scalability, optimizing LLM inference is critical for enhancing efficiency and enabling the deployment of high-performance NLP systems \cite{InstructGPT, polarsparsity}.


Unlike transformer training, which benefits from data parallelism, autoregressive generation remains inherently sequential, limiting scalability. While scaling laws improve performance through larger models and extended training, inference incurs significant computational overhead—generating a single token with a model 10× larger can be 2–3× slower (see Table \ref{tab:model_performance}).

\begin{table}[t!]
\small
    \centering
    \renewcommand{\arraystretch}{1.2}
    \begin{tabular}{lcccc}
        \toprule
        \textbf{Model Name} & \textbf{Milliseconds per token} \\
        \midrule
        meta-llama/Llama-3.2-1B-Instruct & 8.87 \\
        meta-llama/Llama-3.1-8B & 16.65 \\
        meta-llama/Llama-3.1-70B* & 925.05\\
        double7/vicuna-68m & 1.76 \\
        double7/vicuna-160m & 5.61 \\
        lmsys/vicuna-7b-v1.5 &  14.29 \\
        lmsys/vicuna-13b-v1.5 &  20.15 \\
        \bottomrule
    \end{tabular}
    \caption{Inference time comparison. Models marked with * are quantized to int4.} 
    \label{tab:model_performance}
\end{table}


Speculative decoding addresses these inefficiencies by using a smaller "draft" model. This method leverages a smaller "draft" model to generate tokens autoregressively, which are then verified in parallel by the larger "target" model. This increases GPU computations to achieve lower latency. However, this approach introduces a critical hyperparameter, the speculation length ($\gamma$), which represents the number of tokens generated by the draft model before verification.

Selecting an appropriate speculation length is crucial. If $\gamma$ is too large, many tokens generated by the draft model may be rejected, leading to wasted computations and increased latency. Conversely, a small $\gamma$ limits the performance benefits of speculative decoding. Moreover, token generation difficulty varies across sequences—some steps are straightforward and accurately predicted by the draft model, while others require the expertise of the larger target model. Using a constant speculation length throughout generation is suboptimal.


A supervised approach to this challenge involves training a model to adjust speculation length based on prompt and token complexity \cite{dynamic_spec_intel, dynamic_spec_princeton}. While effective, it incurs additional training and computational costs. Instead, we propose an adaptive speculative decoding strategy that dynamically adjusts $\gamma$ in real time via a principled heuristic-driven algorithm, leveraging historical token acceptance to ensure consistent speedups across diverse tasks.

\section{Related Work}

Since the introduction of speculative decoding by \citeauthor{leviathan_kalman_matias_2022} and \citeauthor{chen_borgeaud_irving_lespiau_sifre_jumper_2023}, numerous efforts have been made to enhance its efficiency \cite{liu2024parallelspeculativedecodingadaptive, xiong2024dyspecfasterspeculativedecoding, sun2024spectrfastspeculativedecoding, yang2024multicandidatespeculativedecoding, fastercascadesspeculativedecoding, wertheimer2024acceleratingproductionllmscombined, distillspecimprovingspeculativedecoding, he2024restretrievalbasedspeculativedecoding, bhendawade2024speculativestreamingfastllm}.  

BiLD \cite{kim_mangalam_moon_malik_mahoney_gholami_keutzer_2023} introduces a fallback policy that determines whether to switch to the target model for verification based on the probability of the token generated by the draft model. \citeauthor{spec_infer} and \citeauthor{spectr} propose parallel sampling of tokens from the draft model, constructing draft-token trees that are then verified in parallel by the target model. \citeauthor{jeon_gagrani_goel_park_lee_lott_2024} extend this approach with a recursive speculative decoding algorithm, leveraging the Gumbel trick to sample without replacement and enhance token diversity in the generated draft-token tree. While these methods aim to improve inference speed by increasing the token acceptance rate, they do not guarantee full recovery of the target distribution.  

Other approaches, such as \citeauthor{dynamic_spec_intel} and \citeauthor{dynamic_spec_princeton}, introduce supervised learning techniques to dynamically adjust the speculation length \( \gamma \). They train separate machine learning models to predict when to use the target or draft model for token generation. However, training these models is computationally expensive and highly dependent on the dataset used for inference. As an alternative, \citeauthor{mamou2024accelerating} propose a simple heuristic that adjusts the speculation length based on the number of accepted tokens. While this method avoids the need for additional training, it exhibits high variance in speedup depending on the initial speculation length.  


\section{Background: Speculative Decoding}
\label{background}
\citeauthor{leviathan_kalman_matias_2022} and \citeauthor{chen_borgeaud_irving_lespiau_sifre_jumper_2023} proposed \textbf{speculative sampling} to accelerate inference in large language models by utilizing a smaller \textbf{draft model} to autoregressively sample tokens, while the larger \textbf{target model} verifies these tokens in parallel. This approach has been shown to produce sequences with the same distribution as those sampled directly from the target model. In their setup, the draft and target models differ in size by approximately two orders of magnitude.

Let \( T_{\text{target}} \) denote the time taken by the target model to generate a single token, which is also the time required to verify \( \gamma > 1 \) tokens, assuming sufficient computational resources for parallel processing. Similarly, let \( T_{\text{draft}} \) represent the time taken by the draft model to generate one token.  

We define the \textbf{computational speedup factor} \( c \) as:

\begin{equation}
    c = \frac{T_{\text{target}}}{T_{\text{draft}}}
\end{equation}

Typically, \( c \) ranges from 4 to 10. Given a constant speculation length \( \gamma \) and a goal of generating \( N \) tokens using the target model, our objective is to minimize the total inference cost, defined as:

\begin{equation}
    \text{cost} = T_{\text{target}} \cdot \text{calls}_{\text{target}} + T_{\text{draft}} \cdot \text{calls}_{\text{draft}}
\end{equation}

Let \( \alpha \) be the \textbf{average acceptance rate}—the proportion of draft model tokens accepted by the target model. At each step of speculative decoding, approximately \( \gamma \alpha + 1 \) tokens are accepted. Thus, the total number of decoding iterations required to generate \( N \) target tokens is:

\begin{equation}
    N_{\text{steps}} = \frac{N}{\alpha \gamma + 1}
\end{equation}

The total number of calls to the \textbf{target} and \textbf{draft} models are given by:

\begin{equation}
    \text{calls}_{\text{target}} = \frac{N}{\alpha \gamma + 1}
\end{equation}

\begin{equation}
    \text{calls}_{\text{draft}} = \gamma \cdot \frac{N}{\alpha \gamma + 1}
\end{equation}

Substituting these into the inference cost equation:

\begin{equation}
    \text{cost} = \frac{N}{\alpha \gamma + 1} (c + \gamma) \times T_{draft}
\end{equation}

This equation highlights a trade-off in choosing \( \gamma \):  
\begin{itemize}
    \item Increasing \( \gamma \) reduces the number of calls to the target model, which is desirable.
    \item However, if \( \gamma \) is too large, the acceptance rate \( \alpha \) decreases, leading to an increase in draft model calls and, consequently, a higher total inference cost.
    \item A larger \( c \) allows more draft model calls without significantly increasing cost, emphasizing the importance of selecting an optimal speculation length \( \gamma \) to maintain efficiency.
\end{itemize}

Thus, the choice of \( \gamma \) plays a crucial role in minimizing the overall inference cost.

\section{GammaTune}
\label{gt}
Speculative decoding operates within a dynamically evolving landscape characterized by three distinct regimes based on token acceptance rates as shown in Figure \ref{fig:spec_decode_oracle}. In the easy regime, the draft model remains well-aligned with the target model’s distribution, yielding high acceptance and maximizing parallel decoding efficiency. Conversely, the difficult regime emerges when model distributions diverge, leading to frequent rejections and necessitating near-sequential processing. Between these two extremes lies the moderate regime, wherein the acceptance rate stabilizes around the expected speculative length, striking a balance between acceleration and correction overhead. These regimes are not static; they manifest as a continuum, shifting dynamically based on the interplay between model alignment and decoding progression.

To navigate these regimes adaptively, we introduce GammaTune, an optimization framework that continuously calibrates the speculative decoding window based on an exponentially weighted moving average of historical token acceptance statistics. 


\subsection{Dynamic Adjustment Mechanism}

GammaTune employs a hierarchical control strategy that fuses short-term acceptance signals with long-term statistical adaptation. Let \(\mathcal{A}\) denote the number of accepted tokens in a speculative step. The update mechanism follows:

\paragraph{Adaptive Expansion} If \( \mathcal{A} = \gamma\), an augmentation heuristic increases \(\mathcal{A}\) by a tunable offset \(\delta\), enabling opportunistic window expansion in high-confidence scenarios:

\begin{equation}
    \gamma \gets \mathcal{A} + \delta, \quad \text{if} \quad \mathcal{A} = \gamma.
\end{equation}

\paragraph{Adaptive Window Estimation} The speculative window \(\bar{\gamma}\) is updated via an exponentially weighted moving average while ensuring bounded stability:

\begin{equation}
    \bar{\gamma} \gets \min(\gamma_{\max}, \max(\gamma_{\min}, (1 - \eta) \bar{\gamma} + \eta \mathcal{A})).
\end{equation}

Here, \(\eta\) controls adaptation speed, with lower values enforcing inertia and higher values enabling rapid response.

\begin{algorithm}
\caption{GammaTune Algorithm}
\begin{algorithmic}[1]
\Require $\mathcal{A}, \bar{\gamma}, \gamma, \eta, \gamma_{\min}, \gamma_{\max}, \delta$
\If{$\mathcal{A} = \gamma$}
    \State $\gamma \gets \mathcal{A} + \delta$ \Comment{Increase window by $\delta$}
\EndIf
\State $\bar{\gamma} \gets \min(\gamma_{\max}, \max(\gamma_{\min}, (1 - \eta) \bar{\gamma} + \eta \mathcal{A}))$
\State $\gamma \gets \lceil \bar{\gamma} \rceil$
\end{algorithmic}
\label{alg:gammatune}
\end{algorithm}

This formulation enables GammaTune to adaptively modulate the decoding window—expanding in easy regimes, contracting in difficult ones, and stabilizing in moderate conditions—by seamlessly integrating heuristic adjustments with exponential smoothing to dynamically track evolving token acceptance trends.

\subsection{GammaTune+: Confidence-Guided Early Stopping}
\label{gt_plus}
GammaTune+ enhances GammaTune with a logit-based early stopping criterion. When the draft model’s top logit probability \( p \) falls below a threshold \( \tau \), decoding reverts to sequential verification, adaptively reducing $\gamma$ in low-confidence regions to mitigate inefficiencies while maintaining acceleration in high-certainty regimes.


\section{Experimental Details}

To evaluate our proposed adaptive speculative decoding strategy, we conduct experiments using the SpecBench dataset \cite{xia-etal-2024-unlocking}. This benchmark covers a diverse set of tasks, including writing, roleplay, reasoning, mathematics, coding, information extraction, STEM-related problem-solving, and humanities.

We compare five different speculative decoding methods: \textit{SpecDecode} \cite{leviathan_kalman_matias_2022}, \textit{HFHeuristic} \cite{mamou2024accelerating},  \textit{AssistantThreshold} \cite{mamou2024accelerating}, \textit{GammaTune} (Section \ref{gt}) and \textit{GammaTune+} (Section \ref{gt_plus}). For each method, we conduct experiments using initial speculation lengths of [1, 2, 3, 4, 5, 6, 7, 8, 12, 16, 20, 24] and compute the \textit{average throughput}.


We perform evaluations using the following target/draft model pairs for speculative decoding: Vicuna-13B/Vicuna-160M, Vicuna-7B/Vicuna-68M \cite{vicuna2023}, LLaMA-8B-Instruct/LLaMA-1B-Instruct and LLaMA-70B-Instruct/LLaMA-8B-Instruct \cite{llama3}. All inference experiments are conducted on a single 80GB H100 GPU with KV caching enabled to optimize memory usage and speed.

For all models except LLaMA-70B-Instruct, both the model weights and KV cache are maintained in 16-bit floating-point (float16) precision. Due to its large memory footprint, LLaMA-70B-Instruct is quantized to int8 using the Quanto library from Hugging Face, allowing it to fit within GPU memory constraints while maintaining reasonable performance.


\begin{figure}
    \centering
    \includegraphics[width=0.5\textwidth]{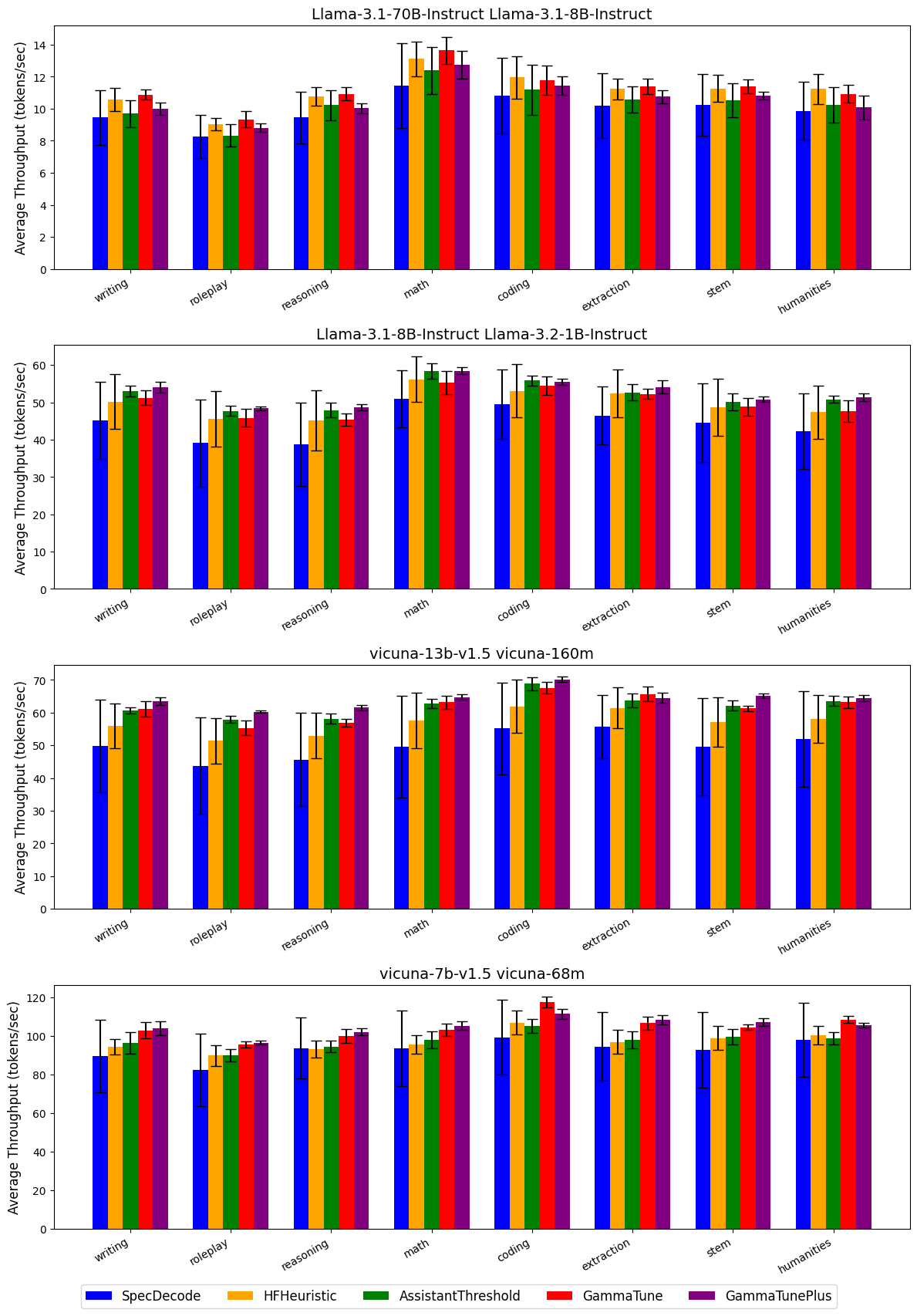}
    \caption{Average Throughput}
    \label{fig:average-throughput}
\end{figure}

\section{Results}

\begin{table}[h]
\centering
\caption{Average speedups over standard speculative decoding. Values after $\pm$ indicate standard deviation across initial $\gamma$ values.}
\resizebox{\columnwidth}{!}{%
  \begin{tabular}{l c c c c c}
    \toprule
    Method
    & \makecell{\texttt{vicuna-13b-v1.5/}\\\texttt{vicuna-160m}}
    & \makecell{\texttt{vicuna-7b-v1.5/}\\\texttt{vicuna-68m}}
    & \makecell{\texttt{Llama-3.1-70B-Instruct/}\\\texttt{Llama-3.1-8B-Instruct}}
    & \makecell{\texttt{Llama-3.1-8B-Instruct/}\\\texttt{Llama-3.2-1B-Instruct}}
    & \textbf{Average} \\
    \midrule
    SpecDecode
      & $1.00 \pm 0.28\times$
      & $1.00 \pm 0.20\times$
      & $1.00 \pm 0.22\times$
      & $1.00 \pm 0.19\times$
      & $1.00 \pm 0.23\times$ \\
    HFHeuristic
      & $1.14 \pm 0.14\times$
      & $1.04 \pm 0.05\times$
      & $1.12 \pm 0.16\times$
      & $1.12 \pm 0.09\times$
      & $1.11 \pm 0.12\times$ \\
    AssistantThreshold
      & $1.24 \pm 0.03\times$
      & $1.05 \pm 0.04\times$
      & $1.04 \pm 0.04\times$
      & $1.17 \pm 0.11\times$
      & $1.13 \pm 0.06\times$ \\
    GammaTune (Ours)
      & $1.23 \pm 0.04\times$
      & $1.13 \pm 0.03\times$
      & $\mathbf{1.13\pm 0.05}\times$
      & $1.12 \pm 0.06\times$
      & $1.15 \pm 0.05\times$ \\
    GammaTune+ (Ours)
      & $\mathbf{1.28 \pm 0.02}\times$
      & $\mathbf{1.13 \pm 0.02}\times$
      & $1.06 \pm 0.02\times$
      & $\mathbf{1.18 \pm 0.05}\times$
      & $\mathbf{1.16 \pm 0.03}\times$ \\
    \bottomrule
  \end{tabular}%
}
\label{tab:speedup_results}
\end{table}

In this section, we compare the performance of different gamma adjustment methods across various tasks in the SpecBench dataset.  Figure \ref{fig:average-throughput} presents a comparison of the \textit{average throughput} (tokens/sec) across four model pairs using various speculative decoding methods. Error bars indicate standard deviations.  For results on acceptance rate and speculation length, refer to Appendix \ref{a1}, and for their relationship to throughput, see Section \ref{background}.

Furthermore, Figure \ref{fig:average-throughput} demonstrates that, on average, \textit{GammaTune} and \textit{GammaTune+} consistently outperform other methods across a range of tasks. Notably, in cases where \textit{GammaTune} underperforms compared to \textit{AssistantThreshold}, \textit{GammaTune+} surpasses both. Moreover, integrating the \textit{AssistantThreshold} heuristic with \textit{GammaTune} in \textit{GammaTune+} helps reduce variance.


Table \ref{tab:speedup_results} demonstrates that GammaTune and GammaTune+ achieve superior average throughput across all tasks, consistently outperforming alternative approaches. This highlights the robustness of our method, which dynamically converges toward near-optimal performance without the need for manual $\gamma$ tuning per dataset. Notably, the low standard deviation across experiments underscores the algorithm’s resilience, ensuring stable and predictable efficiency irrespective of the initial $\gamma$ configuration.



\section{Conclusion}

We proposed a simple, training-free algorithm that consistently outperforms heuristic-based methods and fixed-length speculative decoding across a diverse set of tasks. Our approach demonstrates robust performance with minimal variability, even when the draft and target model latencies differ significantly. Notably, it achieves near-optimal speedups without requiring prior knowledge of the optimal speculation length, making it a reliable choice for speculative decoding in diverse scenarios.

\section{Limitations}


While our approach demonstrates strong performance across SpecBench, its evaluation is limited to four model pairs, primarily from the Vicuna and Llama families. Expanding to a broader range of architectures would enhance generalizability. The benefits of dynamic speculation length ($\gamma$) depend on its variability; when model pairs are well-aligned, low standard deviation might limit adaptive gains. Additionally, reliance on historical token acceptance makes the method susceptible to degradation in volatile or adversarial settings. Future work should explore broader model evaluations, adversarial robustness, and theoretical convergence analysis.

\section*{Acknowledgments}
This research was supported by the National Science Foundation under Grant No. 2203033.

\bibliography{references}

\begin{thebibliography}{26}
\providecommand{\natexlab}[1]{#1}

\bibitem[{Bhendawade et~al.(2024)Bhendawade, Belousova, Fu, Mason, Rastegari, and Najibi}]{bhendawade2024speculativestreamingfastllm}
Nikhil Bhendawade, Irina Belousova, Qichen Fu, Henry Mason, Mohammad Rastegari, and Mahyar Najibi. 2024.
\newblock \href {https://arxiv.org/abs/2402.11131} {Speculative streaming: Fast llm inference without auxiliary models}.
\newblock \emph{Preprint}, arXiv:2402.11131.

\bibitem[{Chen et~al.(2023)Chen, Borgeaud, Irving, Lespiau, Sifre, and Jumper}]{chen_borgeaud_irving_lespiau_sifre_jumper_2023}
Charlie Chen, Sebastian Borgeaud, Geoffrey Irving, Jean-Baptiste Lespiau, Laurent Sifre, and John Jumper. 2023.
\newblock \href {https://arxiv.org/abs/2302.01318} {Accelerating large language model decoding with speculative sampling}.

\bibitem[{Chiang et~al.(2023)Chiang, Li, Lin, Sheng, Wu, Zhang, Zheng, Zhuang, Zhuang, Gonzalez, Stoica, and Xing}]{vicuna2023}
Wei-Lin Chiang, Zhuohan Li, Zi~Lin, Ying Sheng, Zhanghao Wu, Hao Zhang, Lianmin Zheng, Siyuan Zhuang, Yonghao Zhuang, Joseph~E. Gonzalez, Ion Stoica, and Eric~P. Xing. 2023.
\newblock \href {https://lmsys.org/blog/2023-03-30-vicuna/} {Vicuna: An open-source chatbot impressing gpt-4 with 90\%* chatgpt quality}.

\bibitem[{Devlin et~al.(2019)Devlin, Chang, Lee, and Toutanova}]{BERT}
Jacob Devlin, Ming-Wei Chang, Kenton Lee, and Kristina Toutanova. 2019.
\newblock \href {https://doi.org/10.18653/v1/N19-1423} {{BERT}: Pre-training of deep bidirectional transformers for language understanding}.
\newblock In \emph{Proceedings of the 2019 Conference of the North {A}merican Chapter of the Association for Computational Linguistics: Human Language Technologies, Volume 1 (Long and Short Papers)}, pages 4171--4186, Minneapolis, Minnesota. Association for Computational Linguistics.

\bibitem[{Grattafiori et~al.(2024)Grattafiori, Dubey, Jauhri, Pandey, Kadian, Al-Dahle, Letman, Mathur, Schelten, Vaughan, Yang, Fan, Goyal, Hartshorn, Yang, Mitra, Sravankumar, Korenev, Hinsvark, and Rao}]{llama3}
Aaron Grattafiori, Abhimanyu Dubey, Abhinav Jauhri, Abhinav Pandey, Abhishek Kadian, Ahmad Al-Dahle, Aiesha Letman, Akhil Mathur, Alan Schelten, Alex Vaughan, Amy Yang, Angela Fan, Anirudh Goyal, Anthony Hartshorn, Aobo Yang, Archi Mitra, Archie Sravankumar, Artem Korenev, Arthur Hinsvark, and Arun Rao. 2024.
\newblock \href {https://arxiv.org/abs/2407.21783} {The llama 3 herd of models}.

\bibitem[{He et~al.(2024)He, Zhong, Cai, Lee, and He}]{he2024restretrievalbasedspeculativedecoding}
Zhenyu He, Zexuan Zhong, Tianle Cai, Jason~D. Lee, and Di~He. 2024.
\newblock \href {https://arxiv.org/abs/2311.08252} {Rest: Retrieval-based speculative decoding}.
\newblock \emph{Preprint}, arXiv:2311.08252.

\bibitem[{Huang et~al.(2024)Huang, Guo, and Wang}]{dynamic_spec_princeton}
Kaixuan Huang, Xudong Guo, and Mengdi Wang. 2024.
\newblock \href {https://arxiv.org/abs/2405.19715v1} {Specdec++: Boosting speculative decoding via adaptive candidate lengths}.

\bibitem[{Jeon et~al.(2024)Jeon, Gagrani, Goel, Park, Lee, and Lott}]{jeon_gagrani_goel_park_lee_lott_2024}
Wonseok Jeon, Mukul Gagrani, Raghavv Goel, Junyoung Park, Mingu Lee, and Christopher Lott. 2024.
\newblock \href {https://arxiv.org/abs/2402.14160} {Recursive speculative decoding: Accelerating llm inference via sampling without replacement}.

\bibitem[{Kim et~al.(2023)Kim, Mangalam, Moon, Malik, Mahoney, Gholami, and Keutzer}]{kim_mangalam_moon_malik_mahoney_gholami_keutzer_2023}
Sehoon Kim, Karttikeya Mangalam, Suhong Moon, Jitendra Malik, Michael~W Mahoney, Amir Gholami, and Kurt Keutzer. 2023.
\newblock \href {https://arxiv.org/abs/2302.07863} {Speculative decoding with big little decoder}.

\bibitem[{Leviathan et~al.(2022)Leviathan, Kalman, and Matias}]{leviathan_kalman_matias_2022}
Yaniv Leviathan, Matan Kalman, and Yossi Matias. 2022.
\newblock \href {https://arxiv.org/abs/2211.17192} {Fast inference from transformers via speculative decoding}.

\bibitem[{Liu et~al.(2024)Liu, Li, Lv, Liu, Zhu, and Hu}]{liu2024parallelspeculativedecodingadaptive}
Tianyu Liu, Yun Li, Qitan Lv, Kai Liu, Jianchen Zhu, and Winston Hu. 2024.
\newblock \href {https://arxiv.org/abs/2408.11850} {Parallel speculative decoding with adaptive draft length}.
\newblock \emph{Preprint}, arXiv:2408.11850.

\bibitem[{Mamou et~al.(2024{\natexlab{a}})Mamou, Pereg, Korat, Berchansky, Timor, Wasserblat, and Schwartz}]{mamou2024accelerating}
Jonathan Mamou, Oren Pereg, Daniel Korat, Moshe Berchansky, Nadav Timor, Moshe Wasserblat, and Roy Schwartz. 2024{\natexlab{a}}.
\newblock Accelerating speculative decoding using dynamic speculation length.
\newblock \emph{arXiv preprint arXiv:2405.04304}.

\bibitem[{Mamou et~al.(2024{\natexlab{b}})Mamou, Pereg, Korat, Berchansky, Timor, Wasserblat, and Schwartz}]{dynamic_spec_intel}
Jonathan Mamou, Oren Pereg, Daniel Korat, Moshe Berchansky, Nadav Timor, Moshe Wasserblat, and Roy Schwartz. 2024{\natexlab{b}}.
\newblock \href {https://arxiv.org/abs/2405.04304v4} {Dynamic speculation lookahead accelerates speculative decoding of large language models}.

\bibitem[{Miao et~al.(2024)Miao, Oliaro, Zhang, Cheng, Wang, Zhang, Wong, Zhu, Yang, Shi, Shi, Chen, Arfeen, Abhyankar, and Jia}]{spec_infer}
Xupeng Miao, Gabriele Oliaro, Zhihao Zhang, Xinhao Cheng, Zeyu Wang, Zhengxin Zhang, Rae Ying~Yee Wong, Alan Zhu, Lijie Yang, Xiaoxiang Shi, Chunan Shi, Zhuoming Chen, Daiyaan Arfeen, Reyna Abhyankar, and Zhihao Jia. 2024.
\newblock \href {https://doi.org/10.1145/3620666.3651335} {Specinfer: Accelerating large language model serving with tree-based speculative inference and verification}.
\newblock In \emph{Proceedings of the 29th ACM International Conference on Architectural Support for Programming Languages and Operating Systems, Volume 3}, ASPLOS '24, page 932–949, New York, NY, USA. Association for Computing Machinery.

\bibitem[{Narasimhan et~al.(2024)Narasimhan, Jitkrittum, Rawat, Kim, Gupta, Menon, and Kumar}]{fastercascadesspeculativedecoding}
Harikrishna Narasimhan, Wittawat Jitkrittum, Ankit~Singh Rawat, Seungyeon Kim, Neha Gupta, Aditya~Krishna Menon, and Sanjiv Kumar. 2024.
\newblock \href {https://arxiv.org/abs/2405.19261} {Faster cascades via speculative decoding}.
\newblock \emph{Preprint}, arXiv:2405.19261.

\bibitem[{Ouyang et~al.(2022)Ouyang, Wu, Jiang, Almeida, Wainwright, Mishkin, Zhang, Agarwal, Slama, Ray, Schulman, Hilton, Kelton, Miller, Simens, Askell, Welinder, Christiano, Leike, and Lowe}]{InstructGPT}
Long Ouyang, Jeff Wu, Xu~Jiang, Diogo Almeida, Carroll~L. Wainwright, Pamela Mishkin, Chong Zhang, Sandhini Agarwal, Katarina Slama, Alex Ray, John Schulman, Jacob Hilton, Fraser Kelton, Luke Miller, Maddie Simens, Amanda Askell, Peter Welinder, Paul Christiano, Jan Leike, and Ryan Lowe. 2022.
\newblock Training language models to follow instructions with human feedback.
\newblock In \emph{Proceedings of the 36th International Conference on Neural Information Processing Systems}, NIPS '22, Red Hook, NY, USA. Curran Associates Inc.

\bibitem[{Shrestha et~al.(2025{\natexlab{a}})Shrestha, Gautam, and Reddy}]{ssd_espn}
Susav Shrestha, Aayush Gautam, and Narasimha Reddy. 2025{\natexlab{a}}.
\newblock \href {https://doi.org/10.1007/s11227-025-07118-9} {Storage access optimization for efficient gpu-centric information retrieval}.
\newblock \emph{The Journal of Supercomputing}, 81(4):613.

\bibitem[{Shrestha et~al.(2024)Shrestha, Reddy, and Li}]{ESPN}
Susav Shrestha, Narasimha Reddy, and Zongwang Li. 2024.
\newblock \href {https://doi.org/10.1145/3652024.3665515} {Espn: Memory-efficient multi-vector information retrieval}.
\newblock In \emph{Proceedings of the 2024 ACM SIGPLAN International Symposium on Memory Management}, ISMM 2024, page 95–107, New York, NY, USA. Association for Computing Machinery.

\bibitem[{Shrestha et~al.(2025{\natexlab{b}})Shrestha, Settlemyer, Dryden, and Reddy}]{polarsparsity}
Susav Shrestha, Brad Settlemyer, Nikoli Dryden, and Narasimha Reddy. 2025{\natexlab{b}}.
\newblock \href {https://arxiv.org/abs/2505.14884} {Polar sparsity: High throughput batched llm inferencing with scalable contextual sparsity}.
\newblock \emph{Preprint}, arXiv:2505.14884.

\bibitem[{Sun et~al.(2023)Sun, Suresh, Ro, Beirami, Jain, and Yu}]{spectr}
Ziteng Sun, Ananda~Theertha Suresh, Jae~Hun Ro, Ahmad Beirami, Himanshu Jain, and Felix Yu. 2023.
\newblock \href {https://proceedings.neurips.cc/paper_files/paper/2023/file/6034a661584af6c28fd97a6f23e56c0a-Paper-Conference.pdf} {Spectr: Fast speculative decoding via optimal transport}.
\newblock In \emph{Advances in Neural Information Processing Systems}, volume~36, pages 30222--30242. Curran Associates, Inc.

\bibitem[{Sun et~al.(2024)Sun, Suresh, Ro, Beirami, Jain, and Yu}]{sun2024spectrfastspeculativedecoding}
Ziteng Sun, Ananda~Theertha Suresh, Jae~Hun Ro, Ahmad Beirami, Himanshu Jain, and Felix Yu. 2024.
\newblock \href {https://arxiv.org/abs/2310.15141} {Spectr: Fast speculative decoding via optimal transport}.
\newblock \emph{Preprint}, arXiv:2310.15141.

\bibitem[{Wertheimer et~al.(2024)Wertheimer, Rosenkranz, Parnell, Suneja, Ranganathan, Ganti, and Srivatsa}]{wertheimer2024acceleratingproductionllmscombined}
Davis Wertheimer, Joshua Rosenkranz, Thomas Parnell, Sahil Suneja, Pavithra Ranganathan, Raghu Ganti, and Mudhakar Srivatsa. 2024.
\newblock \href {https://arxiv.org/abs/2404.19124} {Accelerating production llms with combined token/embedding speculators}.
\newblock \emph{Preprint}, arXiv:2404.19124.

\bibitem[{Xia et~al.(2024)Xia, Yang, Dong, Wang, Li, Ge, Liu, Li, and Sui}]{xia-etal-2024-unlocking}
Heming Xia, Zhe Yang, Qingxiu Dong, Peiyi Wang, Yongqi Li, Tao Ge, Tianyu Liu, Wenjie Li, and Zhifang Sui. 2024.
\newblock \href {https://doi.org/10.18653/v1/2024.findings-acl.456} {Unlocking efficiency in large language model inference: A comprehensive survey of speculative decoding}.
\newblock In \emph{Findings of the Association for Computational Linguistics ACL 2024}, pages 7655--7671, Bangkok, Thailand and virtual meeting. Association for Computational Linguistics.

\bibitem[{Xiong et~al.(2024)Xiong, Zhang, Li, Wu, and Zou}]{xiong2024dyspecfasterspeculativedecoding}
Yunfan Xiong, Ruoyu Zhang, Yanzeng Li, Tianhao Wu, and Lei Zou. 2024.
\newblock \href {https://arxiv.org/abs/2410.11744} {Dyspec: Faster speculative decoding with dynamic token tree structure}.
\newblock \emph{Preprint}, arXiv:2410.11744.

\bibitem[{Yang et~al.(2024)Yang, Huang, Dai, and Chen}]{yang2024multicandidatespeculativedecoding}
Sen Yang, Shujian Huang, Xinyu Dai, and Jiajun Chen. 2024.
\newblock \href {https://arxiv.org/abs/2401.06706} {Multi-candidate speculative decoding}.
\newblock \emph{Preprint}, arXiv:2401.06706.

\bibitem[{Zhou et~al.(2024)Zhou, Lyu, Rawat, Menon, Rostamizadeh, Kumar, Kagy, and Agarwal}]{distillspecimprovingspeculativedecoding}
Yongchao Zhou, Kaifeng Lyu, Ankit~Singh Rawat, Aditya~Krishna Menon, Afshin Rostamizadeh, Sanjiv Kumar, Jean-François Kagy, and Rishabh Agarwal. 2024.
\newblock \href {https://arxiv.org/abs/2310.08461} {Distillspec: Improving speculative decoding via knowledge distillation}.
\newblock \emph{Preprint}, arXiv:2310.08461.

\end{thebibliography}
\pagebreak
\appendix

\section{Additional Results and Data}
\label{a1}


\begin{figure}[H]
    \centering
    \begin{subfigure}[t]{0.45\textwidth}
        \centering
        \includegraphics[width=\textwidth]{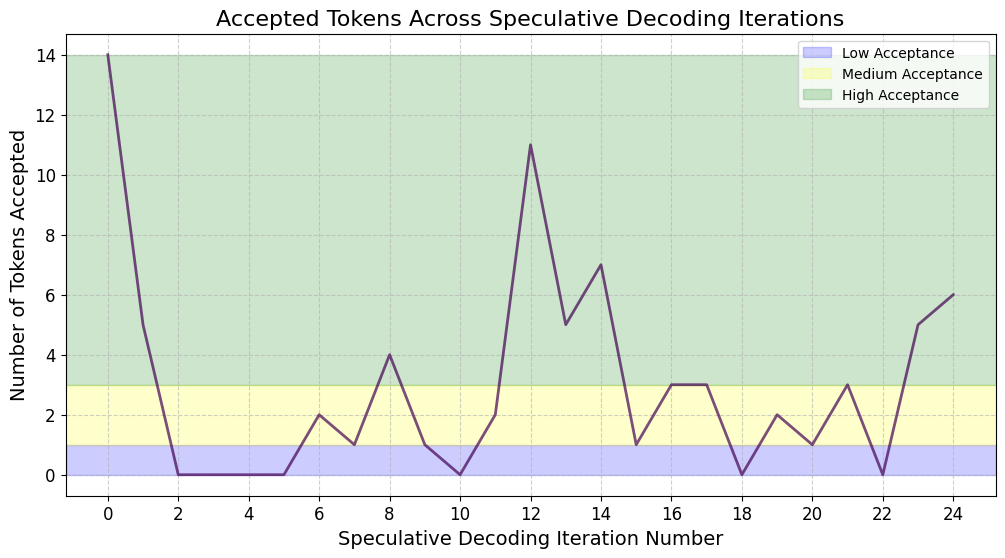}
        \caption{The maximum number of tokens accepted at each step of the speculative decoding process for the text shown in Figure \ref{fig:text_generated}. This illustrates the progression of token acceptance over iterations.}
        \label{fig:spec_decode_oracle}
    \end{subfigure}
    \quad 
    \begin{subfigure}[t]{0.45\textwidth}
        \centering
        \includegraphics[width=\textwidth]{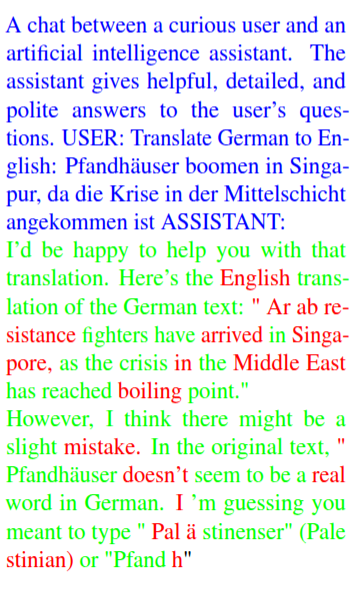}
        \caption{Text generated through speculative decoding using the Llama-70B-Instruct and Llama-8B-Instruct model pair. Tokens in blue represent the prompt, tokens in green are generated by the draft model, and tokens in red are generated by the target model.}
        \label{fig:text_generated}
    \end{subfigure}
    \caption{Visualization of speculative decoding: (a) Maximum number of accepted tokens at each step and (b) Example of generated text using speculative decoding.}
    \label{fig:spec_decode_example}
\end{figure}

\end{document}